\pdfoutput=1

\documentclass[11pt]{article}

\usepackage[]{acl}

\usepackage{times}
\usepackage{latexsym}

\usepackage[T1]{fontenc}

\usepackage[utf8]{inputenc}

\usepackage{microtype}

\usepackage{inconsolata}

\usepackage{makecell}
\usepackage{amssymb}
\usepackage{graphicx}
\title{Enhancing Large Language Model with Decomposed Reasoning for Emotion Cause Pair Extraction}

\author{
    Jialiang Wu \textsuperscript{1} \and
    Yi Shen \textsuperscript{} \and
    Ziheng Zhang \textsuperscript{} \and
    Longjun Cai \textsuperscript{} \\
    \textsuperscript{1} Harbin Institute of Technology \\
        \texttt{\{jialiang.cg,owen.shen.1988,fredzihengzhang,cailongjun\}@gmail.com} 
}

\begin{document}
\maketitle

\graphicspath{{./figs/}}




\begin{abstract}
Emotion-Cause Pair Extraction (ECPE) involves extracting clause pairs representing emotions and their causes in a document. Existing methods tend to overfit spurious correlations, such as positional bias in existing benchmark datasets, rather than capturing semantic features. Inspired by recent work, we explore leveraging large language model (LLM) to address ECPE task without additional training. Despite strong capabilities, LLMs suffer from uncontrollable outputs, resulting in mediocre performance. To address this, we introduce chain-of-thought to mimic human cognitive process and propose the \emph{Decomposed Emotion-Cause Chain (DECC)} framework. Combining inducing inference and logical pruning, DECC guides LLMs to tackle ECPE task. We further enhance the framework by incorporating in-context learning. Experiment results demonstrate the strength of DECC compared to state-of-the-art supervised fine-tuning methods. Finally, we analyze the effectiveness of each component and the robustness of the method in various scenarios, including different LLM bases, rebalanced datasets, and multi-pair extraction.
\end{abstract}

\section{Introduction}

Analyzing the causes behind emotions is an interesting research direction in the field of affective computing, attracting increasing attention in recent years \cite{su2023recent}. Understanding the reason for occurence of emotions holds commercial value in various areas, including product review mining \cite{craciun2019credibility}, social media mining \cite{soong2019essential}, and cognitive-behavioral therapy in psychology \cite{berking2013emotion}.

\begin{figure}[!t]
    \centering
    \includegraphics[width=1\linewidth]{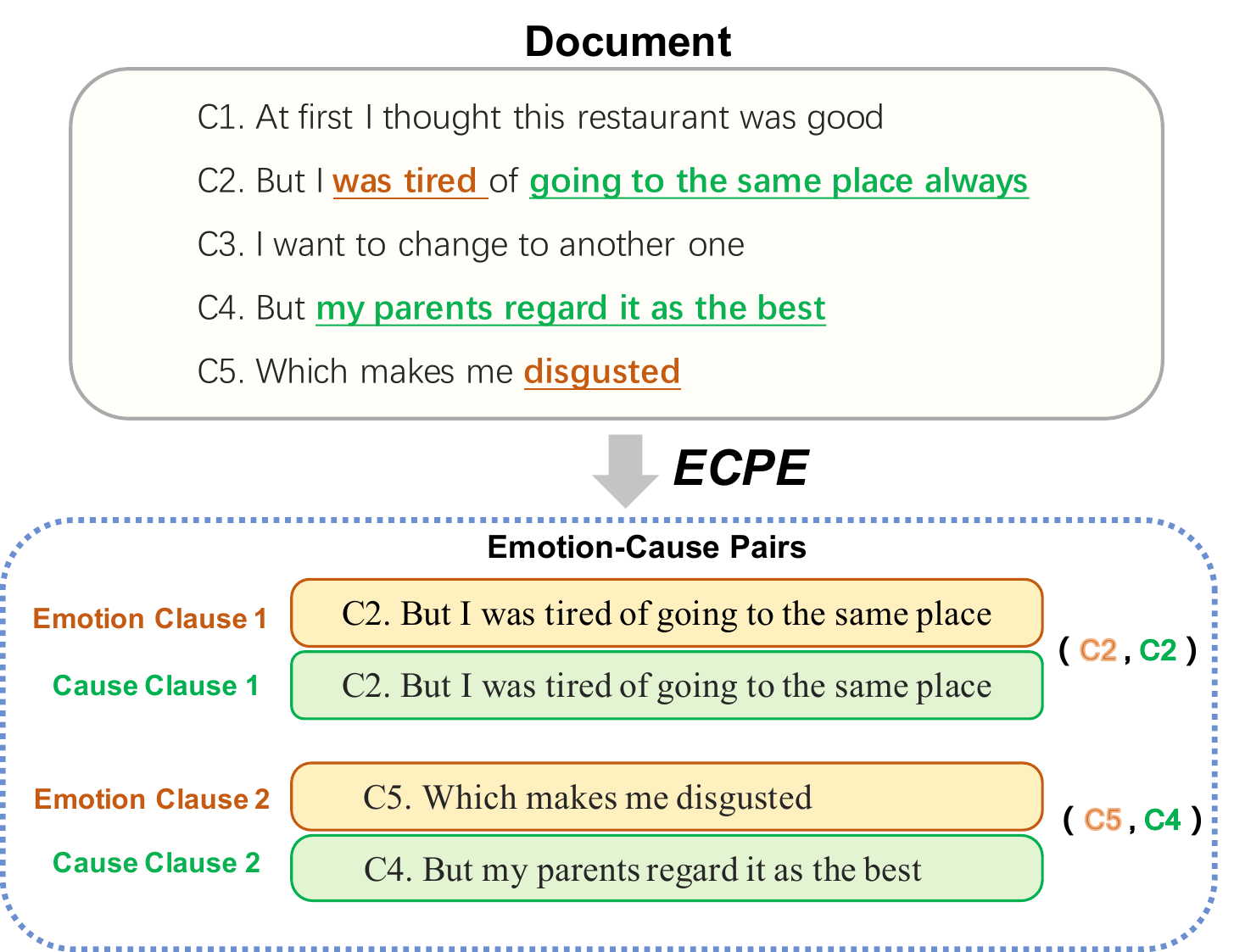}
    \caption{An Example of ECPE task}
    \label{fig:ECPE}
\end{figure}
Early studies on emotion cause analysis primarily focused on clause-level emotion cause extraction (ECE), which aims to extract cause clauses for given emotion clauses. However, the need of emotion annotation limits ECE's application. As a result, the emotion-cause pair extraction (ECPE) task is introduced \cite{002xia2019emotion}. This task aims to extract potential clause pairs of emotions and their corresponding causes in a document. Figure \ref{fig:ECPE} shows an example of ECPE task, where the input document contains several clauses. Both c2 and c5 exhibit emotions. The cause clause for emotion clause c2 is itself, while for emotion clause c5, its associated cause clause is c4. Thus, the output emotion-cause pairs are: (c2, c2), (c5, c4). 

Although researchers have resorted to various deep-learning approaches \cite{su2023recent} to tackle ECPE task, most methods suffer from position bias problem \cite{015zheng2022ueca, 017ding2020experimental}. This issue arises from the relative position information introduced in most existing methods, making ECPE models sensitive to data distribution and lacking robustness. Specifically, given that cause clauses often immediately precede their emotion clause, models may obtain high performance without capturing actual contextual information.

Recently, the rapidly evolving field of natural language processing witnesses the rise of large language model (LLM), such as ChatGPT \footnote{https://chat.openai.com/} and  LLAMA \cite{touvron2023llama}, which have achieved impressive performance in various NLP tasks \cite{sun2023pushing, qin2023chatgpt} under zero-shot or few-shot scenarios, even without updating parameters. Due to their powerful capabilities and significant potential, we make effort to leverage LLMs to address the ECPE task in this paper. 

However, due to the generation paradigm's nature, directly utilizing large models for information extraction tasks often leads to answers with high recall but low precision \cite{han2023information}. In \citet{016wang2023chatgpt}, a constraint is imposed in the prompt, requiring the model to output only one emotion-cause pair. This approach not only fails to achieve satisfactory performance but also conflicts the requirements of practical application scenarios.

Since directly instructing a LLM to output all emotion-cause pairs at once  yields unsatisfactory results, a question arises: `Is there a way to decompose this task and utilize the LLM to get the answer step by step?' Fortunately, the Chain-Of-Thought (COT) technique provides a suitable solution.

The core idea of COT involves generating a series of intermediate reasoning steps, significantly enhancing LLMs' ability to perform complex reasoning. Inspired from existing COT techniques and insights from human cognitive processes \cite{fiske1991social}, we have custom-designed a Decomposed Emotion Cause Chain (DECC) framework. Specifically, We decompose the solving process of the ECPE problem into four steps: \textbf{Recognizing}, \textbf{Locating}, \textbf{Analyzing}, \textbf{Summarizing}. Each of these steps corresponds to a prompt used for interaction with the LLM, and the output from each step becomes part of the prompt for the next step. The entire framework executes in a pipeline manner.

Besides zero-shot decomposed COT, we also employ In-Context Learning (ICL) with automatic selected demonstrations to enhance the framework. Experiment results verify the effectiveness of our method. Our primary contributions are as follows:

\begin{itemize}
    \item We propose a DECC framework to address ECPE task, treating the task as a multi-step reasoning problem. Each step corresponds to a sub-problem tailored to the characteristics of ECPE. The framework then solves the sub-problems sequentially following the decomposed chain-of-thought paradigm.
    \item We employ reliable in-context learning with automatically established demonstrations to further improve the practical performance.
    \item We conduct experiments both on traditional benchmark datasets and improved ones. Our  DECC framework can achieve comparable results to previous supervised fine-tuning methods, including state-of-the-art ones. Further analysis indicates that our approach can effectively alleviate the inherent issues of ECPE that challenge traditional methods.
\end{itemize}

\section{Related Work}

\subsection{Emotion Cause Pair Extraction Task}

After \citet{001chen2010emotion} first introduced the task of Emotion Cause Extraction (ECE),  \citet{002xia2019emotion} proposed Emotion-Cause Pair Extraction (ECPE) to extract emotion clause (EC) and cause clause (CC) simultaneously, which aims to reduce the cost of manual labelling in ECE. Early solutions for the ECPE task are pipeline approaches \cite{002xia2019emotion,013xu2021two}, where EC and CC are extracted sequentially, which will suffer from error propagation issue. Subsequently, many end-to-end approaches \cite{003ding2020end, 004cheng2020symmetric, 006/012wei2020effective, 007chen2020end, 009liu2022pair} have been proposed and achieved better results. Since prompt \cite{014brown2020language} has became popular in the NLP field, \citet{015zheng2022ueca} designed a BERT-based prompting method which achieved state-of-the-art performance. However, \citet{017ding2020experimental} uncovered severe position bias in traditional ECPE benchmark and proposed a rebalanced dataset, on which the performance of most previous carefully designed models decrease drastically.

\subsection{Chain-of-Thought Prompting}

To exploit the reasoning ability in LLMs, \citet{020wei2022chain} proposed Chain-of-Thought (COT) prompting, appending multiple reasoning steps before answering to the input question. With this simple prompting strategy, LLMs are able to perform much better in complex problems. Subsequently, a variety of works \cite{huang2022towards} are proposed to further improve COT prompting in different aspects, including prompt selection \cite{021diao2023active, lu2022dynamic}, self-consistency \cite{026wang2022self}, problem decomposition \cite{khot2022decomposed, zhou2022least}, automatic prompting \cite{023shum2023automatic, 022shao2023synthetic}, and so on.

The most relevant work to this paper is \citet{016wang2023chatgpt}, which evaluates the performance of chatGPT on many tasks, including ECPE. According to its findings,  using chatGPT directly for the ECPE task does not yield satisfactory results. Therefore, we have chosen to enhance models by integrating COT with the task's inherent characteristics.
To the best of our knowledge, this is the first work of leveraging COT in the ECPE task.

\section{Methodology}

\begin{figure*}[ht]
\centering
\includegraphics[width=1.02\textwidth]{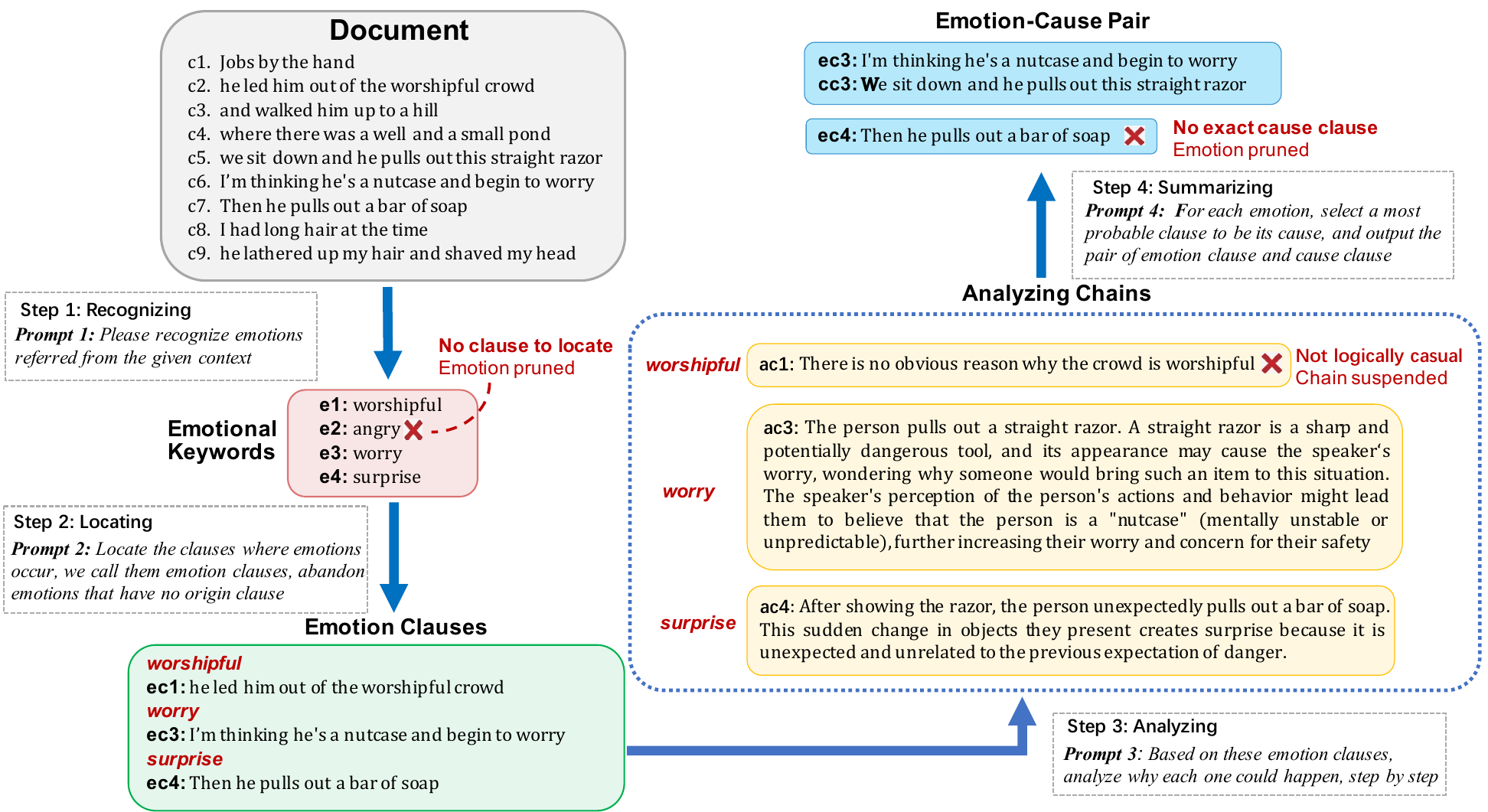} 

\caption{Illustration of DECC framework. At locating step, emotional keyword \textit{angry} is pruned. The analyzing chain corresponding to \textit{worshipful} is pruned at step 3. The analyzing chain of \textit{surprise} is pruned at the last step. \textit{e1-e4} denote the id of specific emotional keyword. \textit{ec} and \textit{cc} denote emotion clause and cause clause, respectively. \textit{ac} stands for the corresponding  analyzing chain.   }
\label{method}
\end{figure*}

\subsection{Task Definition}

In the ECPE task, the input is a document $D$, which consists of several clauses $D=[c_1,c_2,...,c_N]$. Here, $N$ represents the number of clauses in $D$. The objective of the ECPE task is to extract all emotion-cause pairs $({c_e,c_{ca}})$ from $D$ at the clause level, where $c_e$ denotes an emotion clause and $c_{ca}$ represents the corresponding cause clause.

\subsection{DECC Framework}

The Decomposed Emotion-Cause Chain (DECC) is a multi-step reasoning framework based on decomposed chain-of-thought. It involves several interactive prompting steps to optimize the output of large language model. Unlike conventional fine-tuning methods, we aim to utilize the comprehension ability of LLM without additional training cost. As is shown in Figure \ref{method}, we design a series of deducing prompts to filter out improper outputs. Inspired by the human cognitive process and attribution theory \cite{fiske1991social}, which state that when human beings perceive something they are inclined to find the cause, the DECC framework is composed of two major phases: Emotion Extraction and Cause Extraction. 
Each phase also comprises two steps as follows:

\subsubsection{Emotion Extraction Phase}

 The process of emotion extraction consists of the following two steps:

\begin{itemize}

\item \textbf{Step 1. Recognizing}: We instruct LLM with a prompt\footnote{Please refer Figure 2 for details of prompts we use. } to recognize all potential emotions from $D$, presenting them in the form of emotional keywords. It's worth noting that owing to the powerful understanding ability of LLM, the emotions identified here include both explicit and implicit ones. As shown in Figure \ref{method}, the identified emotion \textit{worry} in step 1 belongs to explicit ones (appears in $c6$). Although the emotional keyword \textit{surprise} does not appear in the document, it is distilled by LLM as an implicit emotion based on the content.

\item \textbf{Step 2. Locating}: For each identified emotional keyword, LLM is prompted to find its precise location in document $D$. If a keyword is successfully located, it proceeds to the next step, and its location is considered as a potential emotion clause. On the other hand, if the emotional keyword cannot be located in $D$, it is pruned. As presented in Figure \ref{method}, the emotional keyword \textit{angry} is pruned since there is no clause that explicitly conveys the emotion of anger. For emotional keyword \textit{surprise}, although it is not literally mentioned in $D$, it is implicitly expressed in clause $c7$. In other words, its presence can be inferred from the content of the document.
\end{itemize}
The Emotion Extraction process is expressed by:  
\begin{equation}
    \mathbb{EC}= \{ \textnormal{c} \in C_{loc}\mid\exists e\in\mathrm{E_{rec}}\textnormal{ s.t. } e \subset\textnormal{c} \}
\end{equation}

Where $C_{loc}$ denotes the  clauses located from the passage, $E_{rec}$ denotes the recognized emotions and $\mathbb{EC}$ is the potential  emotion clause set. A clause is eligible to be selected as an emotional clause for the next step only if it contains the recognized emotional keyword; otherwise, it will be pruned. This approach ensures that the locating operation prunes mistakenly recognized emotional keywords, and the recognition step limits the output of clauses.

\subsubsection{Cause Extraction Phase}

When instructing LLM to find a cause clause for each emotion directly, the performance would be disastrous due to the complexity of the ECPE task itself and the uncontrollability of LLM. On the other hand, if we directly ask the LLM to conclude the result pairs, it is very possible for the LLM to obfuscate the granularity of the cause clause. Because every part of reasoning chain could be the cause while only one clause could be the ground truth. Also, because the zero-shot output is unpredictably irregular and contains a large block of tokens, it is challenging to search for final pairs for evaluation. In order to solve the aforementioned problem, we design following two steps:

\begin{itemize}
\item \textbf{Step 3. Analysing}: Given potential emotion clauses ($\mathbb{EC}$), the
 LLM is prompted to analyse the underlying reasons behind each emotion clause in a step-by-step manner. For each emotion clause , we instruct the LLM to generate an analysing chain $ac$, whose beginning is an emotion clause $ec$ and content is the reasoning process based on the overall document $D$. 
 \begin{equation}
     \mathbb{AC}= \{ Analyze\left( ec , D \right) \mid ec\in EC\}
 \end{equation}
 If the reason is successfully deduced and can be attributed to a specific clause in the document, the corresponding rationale (i.e., analyzing chain) generated at this step will be passed to the next. Otherwise, the chain would stop analyzing automatically. As illustrated in Figure \ref{method}, analyzing chain $ac1$ is suspended because $ec1$ is not logically casual according to the document content and has no obvious reason. Consequently, the branch corresponding to $ec1$ and $ac1$ is pruned at this step.

\item \textbf{Step 4. Summarizing}: At this step, the LLM is instructed to identify the most probable cause clause $cc$  for each emotion and summarize final emotion-cause pairs $\left( ec,cc \right) $.
\begin{equation}
    \mathbb{P}=\{ \left( ec,\widehat{cc} \right) \mid \widehat{cc} = argmax P\left(cc \mid ec, ac\right) \}
\end{equation}
Here, some pruning operations will also be carried out. As depicted in Figure \ref{method}, the branch of $ac4$ is pruned  because the cause of \textit{surprise} cannot be simply attributed to a specific clause. Meanwhile, this step is also designed to ensure that the output follows a standardized format, making the generated answer more convenient for further evaluation.
    
\end{itemize}


\subsection{DECC with In-context Learning}

Besides zero-shot prompting, in-context learning (ICL), which incorporates examples within the prompt, is also widely employed in recent works \cite{zhao2023survey} to further enhance larger language models. Since  ICL is very sensitive to the included demonstration examples \cite{dong2022survey}, their selection is a central challenge to address. 

Inspired by previous studies \cite{022shao2023synthetic,fu2022complexity,qiao2022reasoning}, we select demonstrations with the criterion of ensuring diversity. Specifically, we cluster the documents in the training set in a semantic embedding space, using Sentence-BERT \cite{reimers2019sentence} as the encoder. The number of clusters is equal to the number of demonstrations used for inference. We then choose the documents closest to the centroid of each cluster as demonstration candidates. For each candidate document, we follow the DECC framework to generate its rationale step by step. After manually modify these rationales to ensure their accuracy, we put the selected documents along with their corresponding rationales into the prompt, thus completing the construction of the demonstrations for ICL. With selected demonstrations, the LLM will be elicit to generate more intermediate reasoning rationales with high quality which is helpful to the final prediction. More details about ICL-based DECC are elaborated in following sections.

\begin{table*}[!t]
  \centering
  \small
  \renewcommand{\arraystretch}{1.1}
  \begin{tabular}{c ccc ccc |ccc c}
    
    \hline
     & \multicolumn{3}{c}{Chinese dataset} & \multicolumn{3}{c}{English dataset} & \multicolumn{3}{|c}{Rebalanced CN dataset} & De-bias drop\\
     Method & P & R & F1& P & R & F1  & P & R & F1 & F1-ratio\\
    \hline
    ECPE-2D & 72.92 & 65.44 & 68.89 & \textbf{60.49} & 43.84 & 50.73 & 47.22 & 37.38 & 41.73 &-39.43\%\\
    UTOS & 73.89 & 70.62 & 72.03 & 55.69 & 48.03 & \textbf{51.53} & 42.76 & 28.95 & 34.14 & -52.60\%\\
    MTST-ECPE & \textbf{77.46} & 71.99 & 74.63 & 52.37 & 43.54 & 47.47 & 51.99 & 40.34 & 44.93 & -39.80\%\\
    
    RankCP & 71.19 & 76.30 & 73.60 & 44.00 & 45.35 & 44.63 & 43.22 & 39.16 & 41.22 & -43.99\%\\
    ECPE-MLL & 77.00 & 72.35 & 74.52 & 52.96 & 45.30 & 51.21 & \textbf{61.53} & 36.39 & 45.57 & -38.85\%\\
    UECA-Prompt & 71.82 & 77.99 & \textbf{74.70} & - & - & - & 46.30 & 53.22 & 49.37 & -33.91\%\\
    GPT3.5 prompt (0-shot) & 40.74 & 67.54 & 50.82 & 42.11 & 39.34 & 40.68 & 40.72 & 38.14 & 39.39 & -22.49\%\\
    GPT3.5  DECC  (0-shot)& 61.54 & 49.76 & 55.03 & 34.60 & \textbf{59.84} & 43.84 & 45.82 & 49.15 & 47.42 & -13.83\%\\
    GPT3.5  DECC  (4-shot) & 61.23 & \textbf{81.56} & 69.95 & 46.89 & 54.42 & 50.35  & 50.00 & \textbf{79.45} & \textbf{61.38} &  -12.25\%\\
    \hline
  \end{tabular}
  \caption{Results of different methods on three datasets. The results of traditional baselines are from corresponding original papers. “-” means not available in the original paper. The results of GPT3.5 prompt (zero-shot) are reproduced based on the prompt provided in \cite{016wang2023chatgpt}.`De-bias drop' indicates the \textit{F1} reducing ratio of the methods from traditional Chinese dataset to rebalanced Chinese dataset. Best results are marked in \textbf{bold}. }
  \label{tab: main reuslts}
\end{table*}

\section{Experimental Settings}

\subsection{Datasets}

We evaluate our method on the following three published ECPE datasets \footnote{Datasets statistics are listed in Appendix A.}: 

The Chinese ECPE dataset \cite{002xia2019emotion}. It is the most widely used benchmark dataset.

The English ECPE dataset \cite{005singh2021end}. It is mainly composed of an English novel corpus.

The rebalanced Chinese ECPE dataset \cite{017ding2020experimental}. \citet{017ding2020experimental} have identified extreme imbalance of emotion-cause position in the Chinese ECPE dataset. They further uncovered that the performance of current state-of-the-art methods will dramatically degrade without this distribution bias. To address this issue, they proposed this new rebalanced ECPE dataset, which is constructed based on the Chinese ECPE dataset.

Since our method doesn't require any training data, we maintain the same dataset split as previous studies and conduct experiments directly on the test sets \footnote{Implementation details are reported in our Appendix B}.

\subsection{Evaluation Metrics}

We compute the Precision, Recall and F1 score of emotion-cause clause pair for evaluation. However, due to the nature of generative models, evaluation for extraction task such as ECPE is non-trivial \cite{032/gptgen/han2023information}. LLMs often generate reasonable answers that are correct under manual review but fail to match the ground truth \cite{031wadhwa2023revisiting}. Consider the following example, suppose there is a ground truth cause clause \textit{`` my father takes away the toys"}, while the output of LLM might be \textit{``it is author‘s father takes his toys away that most probably triggers the emotion"}. The answer indicates exactly the same information with the target, so it should be regarded as a correct output.
Since the conventional exact matching evaluation for ECPE task is unfair for LLMs, we follow the methods introduced by \citet{016wang2023chatgpt} and conduct human evaluation, enlisting expert annotators to judge whether the model outputs convey the same information as the ground truth.

\subsection{Baselines}

We compare our method with following baselines. \textbf{ECPE-2D} \cite{041ding2020ecpe} is the most outstanding two-step method. End-to-end methods include \textbf{UTOS} \cite{011cheng2021unified}, \textbf{MTST} \cite{042fan2021multi} and \textbf{ECPE-MLL} \cite{003ding2020end} which convert ECPE to a sequence labelling task or multi-label classification task, and \textbf{RankCP} \cite{006/012wei2020effective}  uses BERT+GCN as the clause encoder. \textbf{UECA-Prompt} \cite{015zheng2022ueca} is the state-of-the-art supervised fine-tuning method on both Chinese dataset and rebalanced dataset. We also employ \textbf{GPT3.5} with naive prompting for ECPE task \cite{016wang2023chatgpt} as a LLM baseline.

\section{Results and Analysis}


\subsection{Zero-shot Result}

Since ChatGPT is the most popular LLM, we mainly evaluate DECC based on GPT3.5-turbo in the experiment. As reported in Table \ref{tab: main reuslts}, DECC with GPT3.5-turbo yields a significant improvement, particularly in terms of precision, because many emotion-cause pairs with low probabilities are pruned at step 2 and 3. Despite of some over-pruning operations leads to a certain level of recall degradation, the overall F1 score is still remarkably higher compared to the naive prompting method with GPT3.5-turbo \cite{016wang2023chatgpt}. Nevertheless, DECC under zero-shot settings still lags behind most fully-supervised fine-tuning methods.

\subsection{ICL Few-shot Result}

After enhancing with ICL (4-shot), the performance of DECC further improves. The performance on both Chinese dataset and English dataset are very close to the SOTA fine-tuning based method. Compare to the zero-shot version of DECC, the recall improves significantly and the precision stay stable on Chinese dataset. It seems that GPT3.5-turbo has grasped the specific rule of the task and the impact of over pruning is alleviated after provided with demonstrations. However, the opposite phenomenon occurred on the English dataset, the recall stays stable and the precision improves a lot. We speculate that the reason for this might be due to annotation errors in the dataset. Specifically, for each document in the English ECPE dataset, only one emotion-cause pair is labelled as ground truth target while other potential correction pairs are simply ignored. Therefore, the demonstrations extracted from the English dataset only elicit LLM to generate the most confident outputs, rather than attempting to extract all possible emotion-cause pairs. This makes LLM relatively conservative, resulting in an increase in precision but no improvement in recall.

\subsection{De-bias Result}

In 80\% of the Chinese ECPE dataset samples, emotions and their corresponding causes either appear in adjacent clauses or belong to the same clause, causing a significant position bias \cite{017ding2020experimental}. Therefore, we also examine different methods on the rebalanced dataset. The results indicate that the performance of all previous work decreased drastically by an average of 40\% on F1 score. In comparison, the F1 score decrease of DECC is significantly smaller than other models on the rebalanced dataset and the DECC 4-shot with GPT outperforms the state-of-the-art method with a large margin on F1 score.

Previous methods badly over-fit the spurious correlation introduced by the position bias instead of capturing the semantic information from the context, which causes the deterioration in performance. On the contrary, our approach relies solely on the understanding capability of LLM itself and the reasoning ability provided by the decomposed COT. It is less sensitive to the distribution of dataset. Moreover, we do not build ICL demonstrations specifically for experiment on the rebalanced dataset. Although equipped with the same shots as the normal dataset, DECC still yields an outstanding performance, displaying its excellent generalizing ability.

\subsection{Performance on Multi-pair Document}

There are only less than 30\% documents in the datasets containing more than one emotion-cause pair. Most conventional methods pay little attention to this particular samples, since the performance on multi-pair documents does not affect much on the overall performance. In fact, multi-pair document is very common in practical scenarios and it should not be simply neglected. UECA-Prompt \cite{015zheng2022ueca} is the only existing method which has taken account of multi-pairs extraction and achieved the current state-of-the-art performance. As reported in Table \ref{table:multipair}, DECC 4-shot has surpassed UECA-Prompt with a large margin and achieve new state-of-the-art performance on multi-pair subsets. This result demonstrates that DECC framework is compatible with both multi-pair and single-pair extraction scenarios.

\begin{table}[!h]
  \centering
  \renewcommand{\arraystretch}{1}
  \begin{tabular}{cccc}  
    \hline  
    Methods & P & R & F1 \\
    \cline{1-4}  
       UTOS & 55.45 & 46.76 & 50.35 \\
    RankCP & \textbf{75.08} & 43.90 & 55.31 \\
    ECPE-MLL & 70.45 & 47.76 & 56.88 \\
 
    UECA-Prompt & 69.52 & 54.66 & 61.14 \\
    UECA-Prompt (m2m) & 73.92 & 56.30 & 63.45 \\
    GPT3.5 DECC $_{4-shot}$  & 63.93 & \textbf{73.39} & \textbf{68.34} \\
    \hline  
  \end{tabular}
    \caption{Results on multi-pair extraction scenarios. UECA(m2m) is the special-designed UECA for multi-pair extraction. Best results are marked in \textbf{bold}.}
    \label{table:multipair}
\end{table}
\subsection{Ablation study}
To reveal the specific effect of each step in DECC, we experimented different variants of DECC on GPT3.5 by removing each step respectively \footnote{Necessary changes are added to corresponding prompt for process connection when removing each step}. The experimental results are shown in Table \ref{ablation}, from which we see that all the components make positive contributions to the overall performance. The role of each step is as following:
\begin{itemize}
    \item The performance drop caused by removing the recognizing step indicates its necessity. 
    This is because the subsequent steps introduce many noisy clauses without the initial restriction of emotional keyword in the first step.

    \item Removing the locating step will result in unrelated emotional keywords entering into the analysis step, which might cause explosive emission of reasoning chains. This would produce many redundant pairs in output, which caused the decline of the precision.
    \item DECC suffers a loss of 7.37 F1 score without the analyzing step. The reason is that LLM directly extracts the cause clauses in an end-to-end manner, thereby losing the reasoning ability from COT. 
    \item After removing the summarizing step from DECC, the performance remained almost unchanged. But the output of model will be difficult for human evaluation due to the lack of formatting process.

\end{itemize}
\begin{table}[ht]
  \centering
  \renewcommand{\arraystretch}{1}
  \begin{tabular}{cccc}  
    \hline  
    Methods & P & R & F1  \\
     \hline
     DECC & 61.54 & 49.76 & 55.03  \\
    w/o extracting & 50.41 & 54.46 & 52.36 $\downarrow_{2.67}$  \\
    w/o locating & 47.06 & 53.85 & 50.22 $\downarrow_{4.81}$  \\
 
    w/o analysing & 51.52 & 44.35 & 47.66 $\downarrow_{7.37}$ \\
    w/o summarizing & 59.63 &  49.62 & 54.17 $\downarrow_{0.86}$  \\
    \hline  
  \end{tabular}
    \caption{Ablation experiment results on Chinese dataset}
    \label{ablation}
\end{table}

\subsection{Performance on Different LLMs}

In this subsection, we test our proposed DECC framework on different LLMs. Besides GPT3.5, we also conduct experiments on both Chinese and English datasets using LLAMA \cite{touvron2023llama} and ChatGLM\footnote{https://chatglm.cn/} respectively. As outlined in Table \ref{GLM}, we can see that DECC outperforms naive prompt method significantly across all LLMs with different parameter scales, which illustrates the robustness and universality of our approach.

\begin{table}[!h]
  \centering
  \renewcommand{\arraystretch}{1.03}
  \begin{tabular}{c ccc}  
    \hline  

       & Prompt & DECC & Gain  \\
         \cline{1-4}
    GPT3.5 (Ch.) & 50.82 & 56.73 & +11.63\%  \\
     ChatGLM (Ch.) & 44.73 & 50.51 & +12.92\%\\
     GPT3.5 (En.) & 40.68 & 43.84 & +7.77\% \\
     LLaMA (En.) & 39.72 & 42.55 & +7.12\%\\
    \hline  
  \end{tabular}
    \caption{F1 results of naive prompt and DECC on different LLM bases. `En.' and `Ch.' refer to the English and Chinese dataset. `Gain’ indicates the increase ratio.}
    \label{GLM}
\end{table}

\begin{table*}[!t]
  \centering
  \renewcommand{\arraystretch}{1.12}
  \small
  \begin{tabular}{ ll }  
    \Xhline{1.3pt}
            \textbf{Document:} & \textbf{Document:}\\
     \textit{1. Talking about the scene of parents clearing snow for her} &\textit{1. To prevent Ellie from changing her mind}\\

     \textit{2. Susan's eyes have tears twinkling} &\textit{2. Mike picks her up every day night after school}\\
      
      \textit{3. her parents are more than 60 years old} &\textit{3. for this kind of entanglement}\\
      \textit{4. 30 miles of the mountain road}&\textit{4. Susan had regarded it as a totally trouble} \\
      \textit{5. she said}&\textit{5. so she resolutely broke up with jack} \\
     \textit{6. when she stepped on the back road swept by them}&\textit{6. and asked her father to pick her up every day}\\ 
      \textit{7. the heart is not only moved}  &\textit{7. His girlfriend just broke up with him}\\
      \textit{8. more is regret for not doing filial piety at home }  &\textit{8. this fact made Mike feel very bitter}\\
    \Xhline{0.6pt}
    \textbf{Ground truth:} & \textbf{Ground truth:}\\
        Emotion: \textit{more is regret for not doing filial piety at home} &Emotion: \textit{this fact made Mike feel very bitter}\\
     Cause: \textit{more is regret for not doing filial piety at home} &Cause: \textit{His girlfriend just broke up with him}\\
        Clause Pair: $\left( 8 , 8 \right)$  &Clause Pair: $\left( 8 , 7 \right)$ \\
    \textbf{Output: }   &\textbf{Output: } \\
      \textit{1.Emotion - the heart is not only moved } &\textit{Emotion - made Jack feel very bitter (clause 8)}\\
   \textit{ Reason - parents clearing snow for her}&\textit{Reason - so she resolutely broke up with jack (clause 5)}\\
      \textit{2.Emotion - more is regret for not doing filial piety at home}&(clause 5 and clause 7 both refer to the cause of 'breaking up')\\
    \textit{ Reason - more is regret for not doing filial piety at home}&\\
       
    \Xhline{1.2pt}
  \end{tabular}
    \caption{Two typical label error examples in English dataset. }
    \label{label error}
\end{table*}
\subsection{Scaling the Demonstration Number} 
We investigate the scaling effect of demonstration number for DECC with GPT3.5-turbo. The experiments are conducted on Chinese dataset \cite{002xia2019emotion} and English dataset by varying the demonstrations numbers under the input token limit. We should select the appropriate configurations based on the trade-off between performance and computational cost.
\begin{figure}[htb]
\centering
\includegraphics[width=0.38\textwidth]{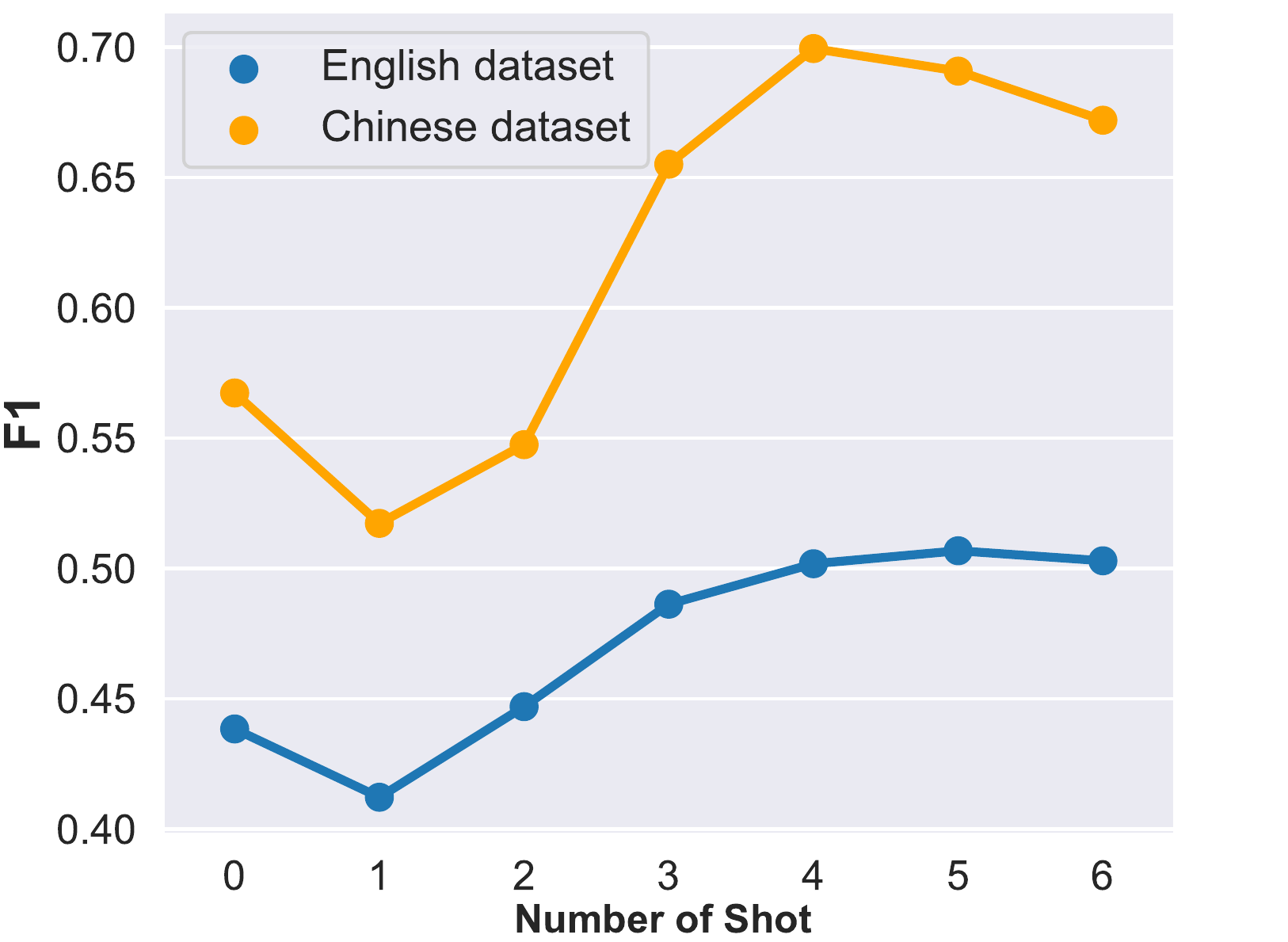}
\caption{Results under different ICL shot numbers on Chinese dataset and English dataset}
\label{fig:lineplot}
\end{figure}

As depicted in Figure \ref{fig:lineplot}, the trends of the two curves are generally consistent. There is a sudden drop of performance at the point of one-shot. A possible reason behind is that LLM could generate output solely based on its comprehending ability under zero-shot settings, while in the ICL one-shot  scenario, LLM might be misled by the only demonstration. For instance, if this demonstration belongs the type of single-pair extraction, then LLM will tend to forcibly extract only one pair for each document, similarly, a multi-pair demonstration may lead to an all-multi-pair extracting phenomenon. As a result, the performance degrades severely. As the shot-number increases, the performance improves significantly and reached the top at four shots. After the shot-number exceeds five, the performance begins to gradually deteriorate. This phenomenon arises may due to the misguidance caused by too many demonstrations, leading the output content to become increasingly unstable.

\subsection{Case Study}


Table \ref{exact match examples} presents a sample from the English dataset, illustrating ground truth labels and two answers generated by DECC 4-shot.  While Output A may not be an exact match, it aligns with the practical requirements of our application. Although Output B does not align with the standard format, it still accurately identifies the cause clause. Therefore, both of them could be considered correct, highlighting the importance of introducing human evaluation.

During the evaluation, we also discover some kinds of label error in the datasets\footnote{Please refer to our Appendix C for more cases.}. Here we display two most common ones, as presented in Table \ref{label error}. In the left column, there are actually two obvious emotion-cause pairs in the document,  whereas the gold truth only contains one pair. LLM with DECC tends to extract both pairs, resulting in an inevitable drop in precision. For the right column, Clause 4 and Clause 7 convey almost identical meanings.  While either could be chosen as the cause clause, in most cases, only the one closer to the emotion clause tends to be labeled as the cause clause. This further highlights how traditional methods can benefit from position bias. These examples illustrate the limitations of the current benchmark datasets. On the other hand, whey also underscore the significant  potential of our proposed DECC. 
\begin{table}[!h]
  \centering
  \renewcommand{\arraystretch}{1.05}
  \small
  \begin{tabular}{l}  
  \Xhline{1.3pt}
     \textbf{Document:}\\
     \textit{1.Hertzfeld took great pride that } \\
      \textit{2.they achieved the functionality solely using software} \\
      \textit{3.he replied} \\
      \textit{4.We don't have any special hardware for it } \\
     \textit{5.Gates insisted special hardware necessary that } \\
      \textit{6.it can move the cursor that way }  \\
      \Xhline{0.4pt}
        \textbf{Ground tructh:}\\
         Emotion clause: \textit{Hertzfeld took great pride that}\\
        Cause clause: \textit{they achieved the functionality using software} \\
                     Clause Pair: $\left( 1 , 2 \right)$  \\
        \textbf{Output A: }\\
    \textit{Emotion clause -``took great pride"}\\
    \textit{Cause clause -``achieve the functionality using software"}\\
    \textbf{Output B: }\\
     \textit{ Emotion: pride (Clause 1) } \\
   \textit{Explanation: The reason for the emotion of pride is the }\\
    \textit{ achievement of being able to achieve their functionality}\\
   \textit{ solely using software, as stated in clause 2}\\
    \Xhline{1.2pt}
  \end{tabular}
    \caption{Examples of LLM producing ``also correct but not exactly matching'' results.}
    \label{exact match examples}
\end{table}
\section{Conclusion}
The primary strategy of DECC framework is to logically break down the ECPE task into a series of subtasks and solve them in sequence. The resolution of each subtask is facilitated by the answers of previous subtasks. With the decomposed chain-of-thought paradigm and meticulously crafted prompts, DECC is able to significantly enhance the reasoning ability of large language models on the ECPE task, while concurrently mitigating challenges faced by traditional methods. Experiments conducted on three benchmark datasets have demonstrated the effectiveness and robustness of DECC. In the future, we aim to explore more efficient and robust demonstration selection methods to further enhance its performance.

\section{Limitations}
Since DECC decomposes the ECPE task into multiple sub-problems, it will increase the inference time of large language models. Besides, we did not test DECC on GPT4 due to the cost constraint.










\bibliography{acl_latex}
\newpage
\appendix

\section{Dataset statistics}
\begin{table}[!h]
  \centering
  \renewcommand{\arraystretch}{1.1}
  \small
  \begin{tabular}{cccc}  
    \hline  
    Statistics  & Chinese & English & Rebalanced  \\
     \hline
     Total documents & 1,945 & 2,843 & 756  \\
    
     One pair doc & 1,746 & 2,537 & 733  \\
 
    Multi-pairs doc & 199 & 306 & 23  \\
    Total pairs & 2,167& 3,215 & 780  \\
    \hline  
  \end{tabular}
    \caption{The statistics of three datasets}
\end{table}

\section{Implementation Details}

For GPT3.5-turbo with naive prompt, we keep exactly the same settings as reported in \cite{016wang2023chatgpt}. For DECC, the tunable hyperparameters include temperature, top-K, penalty factor. These hyperparameters are manually tuned on each LLM with hold-out validation. The results reported in our experiments are all based on the average F1 score of 5 random runs on the test set. During all the experiments. We employed 5 annotators for human evaluation. For each instance, we consider it as correct only when 3 or more annotators agreed that the answer generated by LLM is correct.  

For GPT3.5-turbo and ChatGLM, we obtain corresponding answers through official API. Since LLAMA 2 has not provided an official API, we utilized a third-party API\footnote{https://replicate.com/}. The details about model vision, price and cost are reported in Table \ref{price}.

\begin{table}[ht]
  \centering
  \renewcommand{\arraystretch}{1.1}
  \small
  \begin{tabular}{c|ccccc}  
    \hline  
    Method & Model &  Temperature & Top\_K& Penalty \\
     \hline
      & GPT3.5  & 0 & 1$^{*}$ & 1$^{*}$  \\
    Prompt &ChatGLM & 0 & 1$^{*}$ & - \\
     &LLaMA&   0 & 1$^{*}$ & 1.15$^{*}$  \\
    \hline
      &GPT-3.5& 0.7 & 1 & 0.3  \\
     DECC &ChatGLM & 0.7 & 0.7 & -  \\
    & LLaMA-2  &0.7 & 1 & 0.3   \\

    \hline  
  \end{tabular}
    \caption{Hyperparameters of LLMs in our experiments. ``*" denotes  default setting. ``-" means that ChatGLM API does not provide penalty adjustment. )}
\end{table}

\section{More Case studies }

In some cases, the DECC framework, especially under zero-shot settings,  tends to output underlying causes of emotions rather than direct causes. Table \ref{undelying} shows an example. Although these underlying causes are considered as wrong answer under our human evaluation due to the confliction with the ground truth label, this kind of answers is valuable in many practical application scenarios, which also demonstrates the potential of DECC with LLM. A complete example of DECC workflow is shown in Table \ref{decc}

\begin{table}[!h]
  \centering
  \renewcommand{\arraystretch}{1.15}
  \small
  \begin{tabular}{l}  
  \Xhline{1.3pt}
     \textbf{Document:}\\
     \textit{1.Last week } \\
      \textit{2.outside the burn surgery Department of the Hospital} \\
      \textit{3.the reporter met Mr. Shi} \\
      \textit{4.He regretted the fire in his home} \\
     \textit{5.If I didn't leave the gas tank by the door} \\
      \textit{6.Even if there's an electric fire in the house}  \\
      \textit{7.My wife and my mother will be fine}\\
      \textit{8.And now they are both burned}\\
      \textit{7.It's all my fault}\\
      \Xhline{0.4pt}
        \textbf{Ground tructh:}\\
         Emotion clause: \textit{He regretted the fire in his home}\\
        Cause clause: \textit{He regretted the fire in his home} \\
        Clause Pair: $\left( 4 , 4 \right)$  \\
        \textbf{Output: }\\
    Emotion keyword:  \textit{Regret}\\
    Emotion clause:  \textit{4.He regretted the fire in his home}\\
    Analysis: \textit{Mr.Shi feels regretful to the fire in his house, if }\\
      \textit{  he didn't put the gas tank by the door, the injuries of his}\\
      \textit{  wife and mother wouldn't be like that, so the reason for}\\
      \textit{    regret is putting the gas tank}\\
    Cause clause:  \textit{5.If I didn't leave the gas tank by the door}\\
    \Xhline{1.2pt}
  \end{tabular}
    \caption{Underlying cause}
    \label{undelying}
\end{table}

\begin{table*}[ht]
  \centering
  \renewcommand{\arraystretch}{1.1}
  \small
  \begin{tabular}{cccccc}  
    \hline  
      Model & Version & Provider &  Pricing & Total cost \\
      \hline
      GPT-3.5 Turbo & 175B - July 6 version & OpenAI & USD 0.0015(in) 0.002(out) / 1K tokens& USD 250  \\
     ChatGLM& 130B - STD& ZhiPu & CNY 0.005 / 1K tokens & CNY 32  \\
    LLaMA-2  &70B - chat- HF & Third-Party & USD 0.0032 / second & USD 260  \\

    \hline  
  \end{tabular}
    \caption{Details about LLM API calls.}
    \label{price}
\end{table*}

\begin{table*}[!h]
  \centering
  \renewcommand{\arraystretch}{1.1}
  \small
  \begin{tabular}{l}  
  \Xhline{1.3pt}
     \textbf{Document:}\\
     \textit{1. Mike was angered by this} \\
      \textit{2. after thinking back to the grievance and pain of this period} \\
      \textit{3. He quickly took out the knife from his trouser pocket} \\
      \textit{4. Stabbed Susan four times in the chest} \\
     \textit{5. When Susan's father woke up from his shock} \\
      \textit{6. Mike had run a long way}  \\
      \textit{7. although Susan's father sent her to the hospital as quickly as possible}\\
      \textit{9. Susan suffered a severe chest injury of hemopneumothorax}\\
      \textit{10. because of the deep wound by knife}\\
      
        \textbf{Ground tructh:}\\
         Emotion clause: \textit{When Susan's father woke up from his shock}\\
        Cause clause: \textit{Stabbed Susan four times in the chest} \\
        Clause Pair: $\left( 5 , 4 \right)$  \\
        \Xhline{1.0pt}
        \textit{Subtask 1: }\\
        \textbf{System:}  \\ 
         
         Context : + $[Document]$ + According to the given context, each number at the beginning of line represents a clause, \\
            complete  the following tasks, do not output uncertain answer: \\
        \textbf{User:} \\Please recognize emotions referred from the given context\\
        \textbf{LLM output:}\\ anger, grievance and pain, shock, anxiety \\

\Xhline{0.1pt}
    \textit{Subtask 2: }\\
        \textbf{User:} \\
        Locate the clauses where emotions occur, we call them emotion clauses, abandon  those emotions that have no origin clause\\
        \textbf{LLM output:} \\
        Emotion clauses:\\
        ``Mike was angered by this''\\
        `` after thinking back to the grievance and pain of this period''\\
        `` When Susan's father woke up from his shock''\\

\Xhline{0.1pt}
    \textit{Subtask 3: }\\
        \textbf{User:} \\
        Based on these emotion clauses,
analyze why each one could happen, step by step\\
        \textbf{LLM output:}\\
        1. Emotion: anger\\
         - Mike feel angry may because of certain matters caused by Susan.\\
         2. Emotion: grivance and pain\\
         - Mike thought back to this period, this leads the change of his emotion. \\
         3. Emotion: shock\\
         - Susan's father got shock because of seeing Mike stabbed Susan four times in the chest, this make him very shocked.\\
\Xhline{0.1pt}
        \textit{Subtask 4: }\\
        \textbf{User:} \\
        For each emotion, select a most probable clause to be its cause, and output the pair of emotion clause and cause clause\\
        \textbf{LLM output:} \\
        1. Emotion: anger\\
         - Emotion clause: 1. Mike was angered by this \\
         - None explicitly mentioned in the given context\\
        2. Emotion: grievance and pain \\
         - Emotion clause: 2. after thinking back to the grievance and pain of this period\\
         - None explicitly mentioned in the given context\\
       3. Emotion: shock\\
         - Emotion clause: 5. When Susan's father woke up from his shock\\
         - Cause clause: 4. Stabbed Susan four times in the chest\\ 
    \Xhline{1.2pt}
  \end{tabular}
    \caption{An example of common DECC workflow on LLaMA-2 70B-chat-HF}
    \label{decc}
\end{table*}

\section{The Specific Content of the Prompts}

\begin{figure*}[ht]
    \centering
    \includegraphics[width=0.9\linewidth]{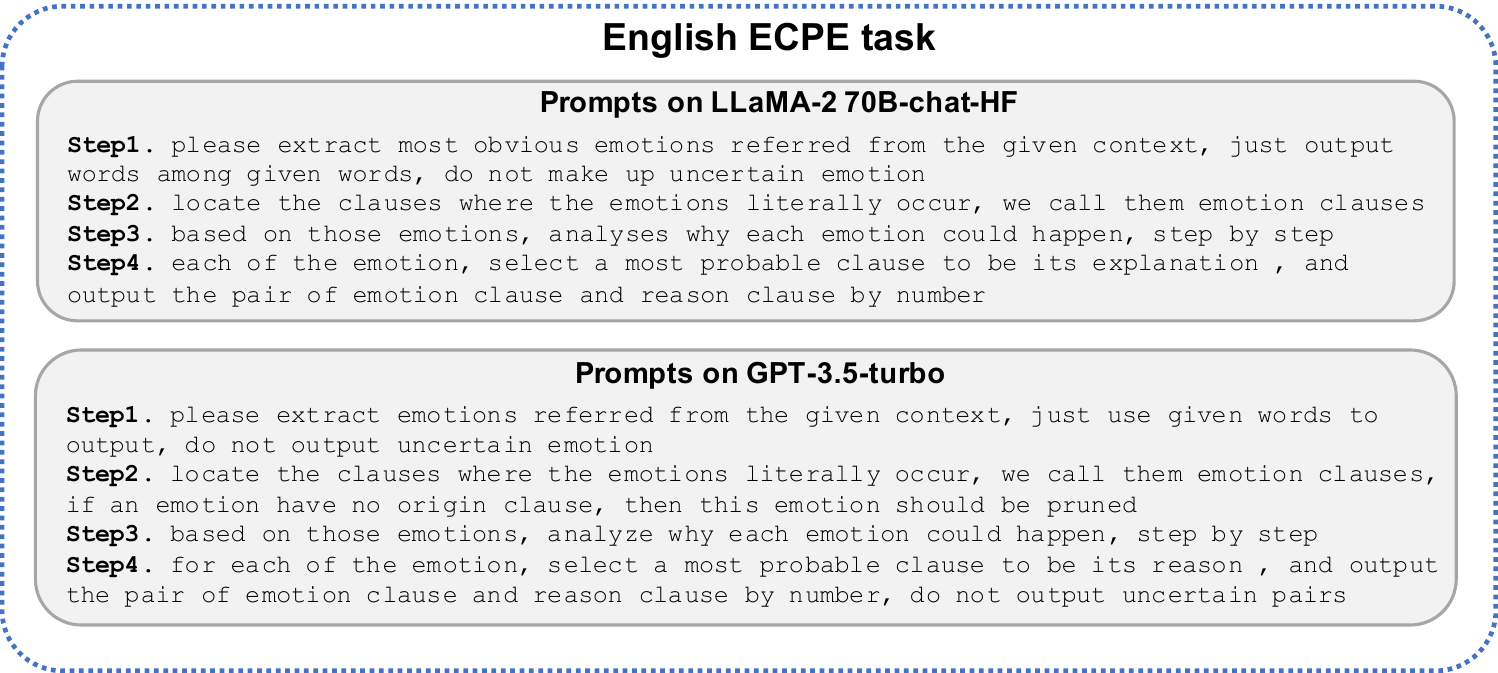}
    \caption{The prompt settings for English dataset.}
    \label{eng}
\end{figure*}

\begin{figure*}[ht]
    \centering
    \includegraphics[width=0.9\linewidth]{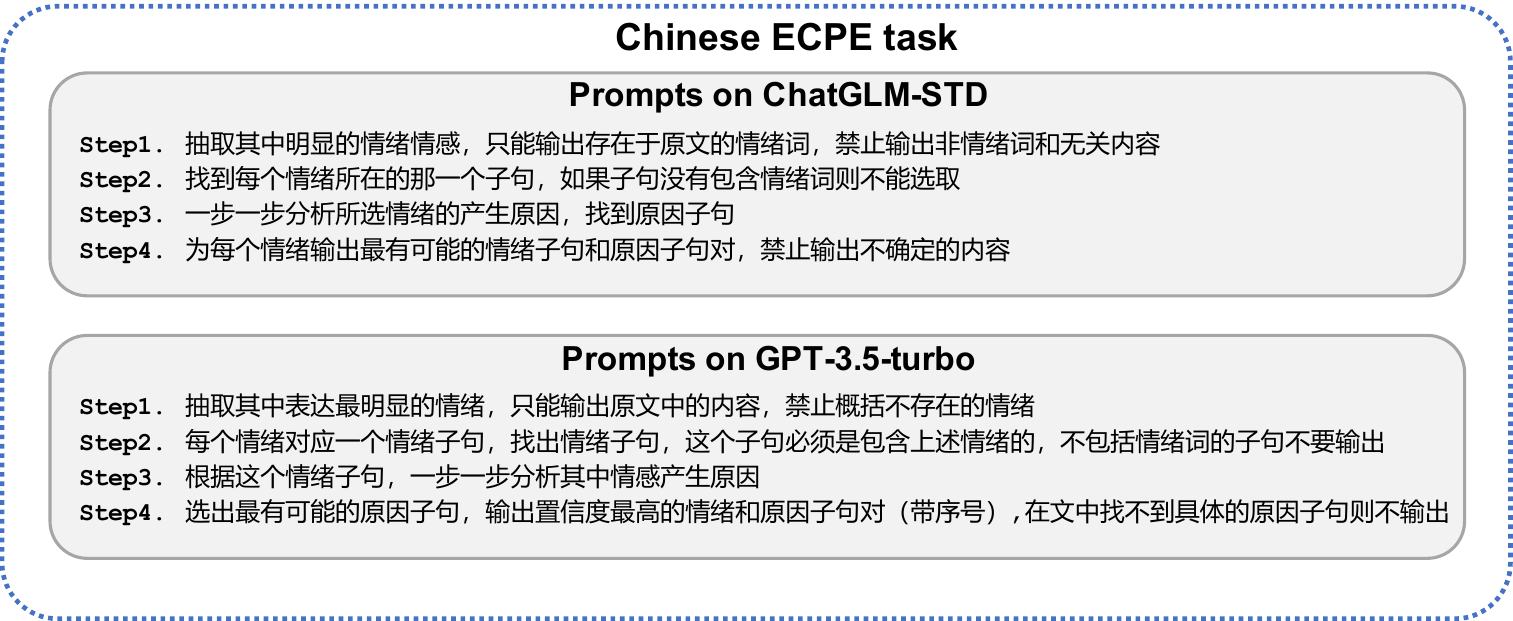}
    \caption{The prompt settings for Chinese dataset.}
    \label{chi}
\end{figure*}

We propose the DECC , aiming at providing a effective multi-stage framework for leveraging LLM for the ECPE task. However, the prompts of different LLMs settings do not have to be strictly consistent. Since LLMs  are pretrained on different corpora and sensitive to different prompts. 
The performance of identical prompts may vary greatly across different LLMs. We provide our English and Chinese prompt settings used in our experiments in Figure\ref{eng} and Figure \ref{chi}, respectively. There are subtle differences between them.



\end{document}